# Self-Organized Stigmergic Document Maps: Environment as a Mechanism for Context Learning

Vitorino Ramos[1] and Juan J. Merelo[2]

*Abstract*— Social insect societies and more specifically ant colonies, are distributed systems that, in spite of the simplicity of their individuals, present a highly structured social organization. As a result of this organization, ant colonies can accomplish complex tasks that in some cases exceed the individual capabilities of a single ant. The study of ant colonies behavior and of their self-organizing capabilities is of interest to knowledge retrieval/ management and decision support systems sciences, because it provides models of distributed adaptive organization which are useful to solve difficult optimization, classification, and distributed control problems, among others. In the present work we overview some models derived from the observation of real ants, emphasizing the role played by stigmergy as distributed communication paradigm, and we present a novel strategy to tackle unsupervised clustering as well as data retrieval problems. The present ant clustering system (ACLUSTER) avoids not only short-term memory based strategies, as well as the use of several artificial ant types (using different speeds), present in some recent approaches. Moreover and according to our knowledge, this is also the first application of ant systems into textual document clustering.

I. STIGMERGY: FROM LOCAL PERCEPTIONS TO GLOBAL ADAPTIVE SOLUTIONS

SYNERGY, from the greek word *synergos*, broadly defined, refers to combined or co-operative effects produced by two or more elements (parts or individuals). The definition is often associated with the quote "the whole is greater than the sum of its parts" (Aristotle, in *Metaphysics*), even if it is more accurate to say that the functional effects produced by wholes are different from what the parts can produce alone [5]. Synergy is a ubiquitous phenomenon in nature and human societies alike. One well know example is provided by the emergence of self-organization in social insects, via direct (mandibular, antennation, chemical or visual contact, etc) or indirect interactions. The latter types are more subtle and defined by *Grassé* as stigmergy [10,11] to explain task coordination and regulation in the context of nest reconstruction in *Macrotermes* termites. An example [1], could be provided by two individuals, who interact indirectly when one of them modifies the environment and the other responds to the new environment at a later time. In other words, stigmergy could be defined as a typical case of environmental synergy. *Grassé* showed that the coordination and regulation of building activities do not depend on the workers themselves but are mainly achieved by the nest structure: a stimulating configuration triggers the response of a termite worker, transforming the configuration into another configuration that may trigger in turn another (possibly different) action performed by the same termite or any other worker in the colony. Another illustration of how stimergy and self-organization can be combined into more subtle adaptive behaviors is recruitment in social insects, as in nest cleaning by some workers [1].

Division of labor is another paradigmatic phenomenon of stigmergy. Simultaneous task performance (parallelism) by specialized workers is believed to be more efficient than sequential task performance by unspecialized workers [12]. Parallelism avoids task switching, which costs energy and time. A key feature of division of labor is its plasticity [21]. Division of labor is rarely rigid. The ratios of workers performing the different tasks that maintain the colony's viability and reproductive success can vary in response to internal perturbations or external challenges. But by far more crucial to the present work and aim, is how ants form piles of items such as dead bodies (corpses), larvae, or grains of sand (fig. 1). There again, stigmergy is at work: ants deposit items at initially random locations. When other ants perceive deposited items, they are stimulated to deposit items next to them, being this type of cemetery clustering organization and brood sorting a type of self-organization and adaptive behavior. There are other types of examples (e.g. prey collectively transport), yet stimergy is also present: ants change the perceived environment of other ants (their cognitive map, according to *Chialvo* and *Millonas* [3,15,16]), and in every example, the environment serves as medium of communication [1].

Nevertheless, what all these examples have in common is that they show how stigmergy can easily be made operational. As mentioned by *Bonabeau* et al. [1], that is a promising first step to design groups of artificial agents which solve problems: replacing coordination (and possible some hierarchy) through direct communications by indirect interactions is appealing if one wishes to design simple agents and reduce communication among agents. Finally, stigmergy is often associated with flexibility: when the environment changes because of an external perturbation, the insects respond *appropriately* to that perturbation, as if it were a modification of the environment caused by the colony's activities. In other words, the colony can collectively respond to the perturbation with individuals exhibiting the same behavior. When it comes to artificial agents, this type of flexibility is priceless: it means that the agents can respond to a perturbation without being reprogrammed to deal with that particular instability. In our context, this means that no classifier re-training is needed for any new sets of data-item types (new classes) arriving to the system, as is necessary in many classical models, or even in some recent ones. Moreover, the data-items that were used for supervised purposes in early stages in the colony evolution in his exploration of the search-space, can now, along with new items, be re-arranged in more optimal ways. Classification and/or data retrieval remains the same, but the system organizes itself in order to deal with new classes, or even new sub-classes. This task can be performed in real time, and in robust ways due to system's redundancy. Recently, several papers have highlighted the efficiency of stochastic approaches based on ant colonies for problem solving. This concerns for instance combinatorial optimization problems like the Traveling Salesman problem, the Quadratic Assignment problem, Routing problem, the Bin Packing problem, or Time Tabling problems. Numerical optimization problems have been tackled also with artificial ants, as well as Robotics.

[1] CVRM-IST Geo-Systems Center, Instituto Superior Técnico, Av. Rovisco Pais, 1049-001, Lisbon, PORTUGAL, vitorino.ramos@alfa.ist.utl.pt.
[2] Grupo GeNeura, Dpto. de Arquitectura y Tecnología de Computadores, Facultad de Ciencias, Campus Fuentenueva, s/n 18071, Granada, SPAIN, jmerelo@geneura.ugr.es.

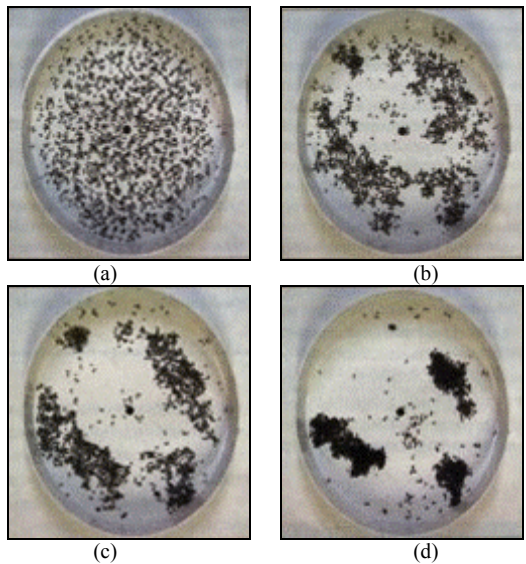

Fig. 1. From (a) to (d), a sequential clustering task of corpses performed by a real ant colony. 1500 corpses are randomly located in a circular arena with radius = 25 cm, where *Messor Sancta* workers are present. The figure shows the initial state (a), 2 hours (b), 6 hours (c) and 26 hours (d) after the beginning of the experiment (from [4]).

Data clustering is also one of those problems in which real ants can suggest very interesting heuristics for computer scientists. One of the first studies using the metaphor of ant colonies related to the above clustering domain is due to *Deneubourg* [7], where a population of ant-like agents randomly moving onto a 2D grid are allowed to move basic objects so as to cluster them. This method was then further generalized by *Lumer* and *Faieta* [14], applying it to exploratory data analysis, for the first time. In 1995, the two authors were then beyond the simple example, and applied their algorithm to interactive exploratory database analysis, where a human observer can probe the contents of each represented point (sample, image, item) and alter the characteristics of the clusters. They showed that their model provides a way of exploring complex information spaces, such as document or relational databases, because it allows information access based on exploration from various perspectives. However, this last work entitled "Exploratory Database Analysis via Self-Organization", according to [1], was never published due to commercial applications. They applied the algorithm to a database containing the "profiles" of 1650 bank customers. Attributes of the profiles included marital status, gender, residential status, age, a list of banking services used by the customer, etc. Given the variety of attributes, some of them qualitative and others quantitative, they had to define several dissimilarity measures for the different classes of attributes, and to combine them into a global dissimilarity measure (in, pp. 163, Chapter 4 [1]). Our aim is to improve these models, introducing some radical changes and different ant-like heuristics, developing a model without any local memory and/or hybridization with more classical approaches. Moreover, the present work will be applied for the first time to document filtering and document exploratory data analysis. The datasets represent a collection of 931 words extracted from a Spanish newspaper. But let us first review some models.

## II. CORPSE CLUSTERING, BROOD SORTING MODELS AND VARIANTS INTO EXPLORATORY DATA ANALYSIS

In several species of ants, workers have been reported to sort their larvae or form piles of corpses – literally cemeteries – to clean up their nests. *Chrétien* [32] has performed experiments with the ant *Lasius niger* to study the organization of cemeteries. Other experiments include the ants *Pheidole pallidula* reported in [7] by *Denebourg* et al., and many species actually organize a cemetery. Figure 1 (section I) shows the dynamics of cemetery organization in another species: *Messor sancta*. If corpses, or more precisely, sufficiently large parts of corposes ara randomly distributed in space at the beginning of the experiment, the workers form cemetery clusters within a few hours, following a behavior similar to aggregation. If the experimental arena is not sufficiently large, or if it contains spatial heterogeneities, the clusters will be formed along the edges of the arena or, more generally, following the heterogeneities. The basic mechanism underlying this type of aggregation phenomenon is an attraction between dead items mediated by the ant workers: small clusters of items grow by attracting workers to deposit more items. It is this positive and auto-catalytic feedback that leads to the formation of larger an larger clusters. In this case, it is therefore the distribution of the clusters in the environment that plays the role of stigmergic variable.

*Denebourg* et al. [7] have proposed two closely related models to account for the two above-mentioned phenomenon of corpse clustering and larval sorting in ants. Although the model of clustering reproduces experimental observations more faithfully, the second one gives rise to more applications. Both models rely on the same principle, and, in fact, the clustering model is merely a special case of the sorting model. The general idea is that isolated items should be picked up and dropped at some other location where more items of that type are present. Let us assume that there is only one type of item in the environment. The probability $P_p$ for a randomly moving, unladen agent (representing an ant in the model) to pick up an item is given by:

$$P_p = \left( \frac{k_1}{k_1 + f} \right)^2 \quad (2.1)$$

where $f$ is the perceived fraction of items in the neighborhood of the agent, and $k_1$ is a threshold constant. When $f \ll k_1$, $P_p$ is close to 1, that is, the probability of picking up an item is high when there are not many items in the neighborhood. $P_p$ is close to 0 when $f \gg k_1$, that is, items are unlikely to be removed from dense clusters. The probability $P_d$ for a randomly moving loaded agent to deposit an item is given by:

$$P_d = \left( \frac{f}{k_2 + f} \right)^2 \quad (2.2)$$

where $k_2$ is another threshold constant: for $f \ll k_2$, $P_d$ is close to 0, whereas for $f \gg k_2$, $P_d$ is close to 1. In their simulations, *Denebourg* et al. [22] have used $k_1 = 0.1$ and $k_2 = 0.3$, testing the spatial sorting organization of 400 items of two types, on a 100 x 100 grid, using 10 agents and $T = 50$; 5,000,000 iterations were needed to accomplish a feasible visual result. As expected, the depositing behavior obeys roughly opposite rules. In order to evaluate $f$, *Denebourg* et al., having a robotic implementation in mind, assumed that $f$ is computed through a short-term memory that each agent possesses: an agent keeps track of the last $T$ time units, being $f$ simply the number $N$ of items encountered during these last $T$ time units, divided by the largest possible number of items that can be encountered during $T$ time units. If one assumes that only 0 or 1 object can be found within a time unit, then $f=N/T$. Their simulations [7], show how small evenly spaced clusters emerge within a relatively short time and then merge, more slowly, into fewer larger clusters, achieving a spatial distribution of objects very similar to those found in nature (fig. 1).

But, as mentioned above, this procedure lends itself more easily to a robotic implementation. As we shall see, the algorithms later described (as well as those proposed) in the present work, are inspired by this idea, but rely on a more direct evaluation of $f$. This procedure should, therefore, be taken as an example among many possible procedures, and changing the detail how $f$ is perceived does not drastically alter the results, according to *Bonabeau* [2]. Among other differences proposed later, are also those directly related to how the agents move on the spatial grid. For instance, real ants are likely to use chemical or tactile cues to orient their behavior. In their simulations, however, *Denebourg* et al. [7] have taken the option of using randomly moving agents, while in here and due to our aim, we suggest the use of ant-like spatial transition probabilities (section III), based on chemical pheromone non-linear weighting functions.

In order to consider sorting, let us now assume that there are two types, $A$ and $B$, of items present in the environment. The principle is the same as before, but now $f$ is replaced by $f_A$ and $f_B$, the respective fractions of items of types $A$ and $B$ encountered during the last $T$ time units. Even if several applications could be derived from here (e.g., segregation phenomenon: *Melhuish* et al., [1], pp. 178), the model is unable to reproduce exactly the brood sorting patterns observed in *Leptothorax* ants, where brood sorting is organized into concentric areas of different brood types.

Significantly more interesting to the present proposal is however, *Lumer's* and *Faieta* model [14]. Both authors have generalized *Denebourg* et al.'s [7] BM to apply it to exploratory data analysis. The idea is to define a distance or dissimilarity $d$ between objects in the space of object attributes. For instance, in the BM, two objects $o_i$ and $o_j$ can only be either similar or different, so that a binary distance can be defined, where, for example, $d(o_i, o_j) = 0$ if $o_i$ and $o_j$ are identical objects, and $d(o_i, o_j) = 1$ if $o_i$ and $o_j$ are not identical objects. Obviously, the very same idea can be extended to include more complicated objects, that is, objects with more attributes, and/or more complicated distances. It is classical in data analysis to have to deal with objects that can be described by a finite number $n$ of real-valued attributes (features), so that objects can be seen as points in $R^n$, and $d(o_i, o_j)$ is the euclidean norm, or any other usual metric, such as the infinite norm $\|...\|_\infty$ or the *Mahalanobis* metric.

The algorithm introduced by *Lumer* and *Faieta* [14] (hereafter LF) consists of projecting the space of attributes onto some lower dimensional space, typically of dimension $z = 2$, so as to make clusters appear with the following property: intra-cluster distances (i.e., attribute distances between objects within clusters) should be small with respect to inter-clusters distances, that is, attribute distances between objects that belong to different clusters. Such a mapping should, therefore, keep some of the neighborhood relationships present in the higher-dimensional space (which is relatively easy since, for instance, any continuous mapping can do the job) without creating too many new neighbors in $m$ dimensions, $m < n$, that would be false neighbors in $n$ dimensions (which is much less trivial since projections tend to compress information and may map several well-separated points in the $n$-dimensional space onto one single point in the $m$-dimensional space). Now, the LF algorithm works as follows. Let us assume that $z = 2$. Instead of embedding the set of objects into $R^2$ or a subspace of $R^2$, they approximate this embedding by considering a grid, that is, a subspace of $Z^2$, which can also be considered a discretization of a real space. Ants that are moving in this discrete space can directly perceive a surrounding region of area $s^2$ (a square $Neigh_{(s \times s)}$ of $s \times s$ sites surrounding site $r$). Direct perception allows a more efficient evaluation of the state of the neighborhood than the memory-based procedure used in the BM algorithm: while the BM was aimed to a robotic implementation, the LF algorithm is to be implemented in a computer, with significantly fewer material constraints. Let $d(o_i, o_j)$ be the distance between two objects $o_i$ and $o_j$ in the feature space. Let us also assume that an agent is located at site $r$ at time $t$, and finds an object $o_i$ at that site. The "local density" $f(o_i)$ with respect to object $o_i$ at site $r$ is given by (Eq. 2.3):

$$f(o_i) = \max\left\{0, \frac{1}{s^2} \sum_{o_j \in Neigh_{(sxs)}(r)} \left[1 - \frac{d(o_i, o_j)}{\alpha}\right]\right\}$$

$f(o_i)$ is a measure of the average similarity of object $o_i$ with the other objects $o_j$ present in the neighborhood of $o_i$. That is, $f(o_i)$ replaces the fraction $f$ of similar objects in the BM model, while $\alpha$ is a factor that defines the scale of dissimilarity: it is important for it determines when two items should or should not be located next to each other. For example, if $\alpha$ is too large, there is not enough discrimination between different items, leading to the formation of clusters composed of items which should not belong to the same cluster. If, on the other hand, $\alpha$ is too small, distances between items in attribute space are amplified to the point where items which are relatively close in attribute space cannot be clustered together because discrimination is too high. Then, and inspired by *Denebourg* et al.'s functions [7] (Eqs. 2.1 and 2.2), *Lumer* and *Faieta* [14] defined picking up and dropping probabilities as follows:

$$P_p(o_i) = \left(\frac{k_1}{k_1 + f(o_i)}\right)^2 \quad (2.4)$$

$$P_d(o_i) = 2f(o_i), \quad if \quad f(o_i) < k_2$$
$$P_d(o_i) = 1, \quad if \quad f(o_i) \geq k_2 \quad (2.5)$$

where $k_1$ and $k_2$ are two constants that play a role similar to $k_1$ and $k_2$ in the BM. *Lumer* and *Faieta* [14] have used $k_1 = 0.1$, $k_2 = 0.15$ (while BM uses $k_2 = 0.3$) and $\alpha = 0.5$, with $t_{max} = 10^6$ steps. In order to illustrate the functioning of their algorithm, the authors used a simple example in which the attribute space is $R^2$, and the values of the two attributes for each object correspond to its coordinates $(x,y)$ in $R^2$. Four clusters of 200 points each were generated in attribute space, with $x$ and $y$ distributed according to Normal (or Gaussian) distributions $N(\mu,\sigma)$ of average $\mu$ and variance $\sigma^2$ (the same distribution was later used for tests in the present work – figure 2, section V). The data points (items) were then assigned at random locations on a 100 x 100 non-toroidal grid, and the clustering algorithm was run with 10 ants. As a result, objects that are clustered together belong generally to the same initial distribution, and objects that do not belong to the same initial distribution are found generally in different clusters.

*Lumer* and *Faieta* [14] have then added three features to their system, due to the fact that are generally more clusters in the projected system than in the initial distribution. These features help to solve this problem, even if they are computationally intensive and broadly bio-inspired. They are:

• *Ants with different moving speeds*. The swarm is distributed uniformly in the interval $[1, v_{max}]$ of possible speed behaviors, where $v$ is the number of grid units walked per time unit by an ant along a given grid axis ($v_{max} = 6$). The speed $v$ influences the tendency of an ant to either pick-up or drop an object through a $f(o_i)$ function similar to Eq. 2.3, where $\alpha$ is replaced by the term $\alpha(1+((v-1)/v_{max}))$. That is, fast moving ants are not as selective as slow ants in their estimation of the average similarity of an object to its neighbors. The diversity of ants allows to form clusters over various scales simultaneously: fast ants form coarse clusters on large scales, i.e. drop items approximately in the right coarse grained region, while slow ants take over at smaller scales by placing objects with more accuracy.

• *A short term memory*. Ants can remember the last $m$ items they have dropped along with their locations. Each time an item is picked up, the ant compares the properties of the item with those of the $m$ memorized items and goes towards the location of the most similar item instead of moving randomly. This behavior leads to a reduction in the number of statistically equivalent clusters, since similar items have a lower probability of initiating independent clusters, as argued in [14].

• *Behavioral switches*. The system exhibits some kind of self-annealing global behavior since items are less and less likely to be manipulated as clusters of similar objects from. Both authors have added the possibility for agents to start destroying clusters if they have not performed any deposit or pick up actions for a given number of time steps. This procedure allows a "heating up" of the system to escape local non-optimal spatial configurations.

Finally, *Lumer* and *Faieta* [14] suggest that their algorithm is halfway between a cluster analysis – insofar as elements belonging to different concentration areas in their *n*-dimensional space end-up in different clusters – and a multi-dimensional scaling, in which an intra-cluster structure is constructed. Note that in the present example, the exact locations of the various clusters on the two-dimensional space are arbitrary, whereas they usually have a meaning in classical factorial analysis. As mentioned by *Dorigo* et al. [8], in a lot of cases, information about the locations of the clusters is not necessary or useful (especially in the context of textual databases) and relaxing the global positioning constraints allows to speed-up the clustering process significantly.

III. FROM RANDOMLY MOVING AGENTS TO BIO-INSPIRED SPATIAL PROBABILITIES

Instead of trying to solve some disparities in the basic LF algorithm by adding different ant casts, short-term memories and behavioral switches (described in section II) which are computationally intensive, representing simultaneously a potential and difficult complex parameter tuning, it is our intention (within the present ACLUSTER proposal) to follow real ant-like behaviors as possible (some other features will be incorporated, as the use of different response thresholds to task-associated stimulus intensities, discussed later at section IV). In that sense, bio-inspired spatial transition probabilities are incorporated into the system, avoiding randomly moving agents, which tend the distributed algorithm to explore regions manifestly without interest (e.g., regions without any type of object clusters), being generally, this type of exploration, counterproductive and time consuming. Since this type of transition probabilities depend on the spatial distribution of pheromone across the environment, the behavior reproduced is also a stigmergic one. Moreover, the strategy not only allows to guide ants to find clusters of objects in an adaptive way (if, by any reason, one cluster disappears, pheromone tends to evaporate on that location), as the use of embodied short-term memories is avoided (since this transition probabilities tends also to increase pheromone in specific locations, where more objects are present). As we shall see, the distribution of the pheromone represents the memory of the recent history of the swarm, and in a sense it contains information which the individual ants are unable to hold or transmit. There is no direct communication between the organisms but a type of indirect communication through the pheromonal field. In fact, ants are not allowed to have any memory and the individual's spatial knowledge is restricted to local information about the whole colony pheromone density. In order to design this behavior, one simple model was adopted (*Chialvo* and *Millonas*, [3]), and extended due to specific constraints of the present proposal.

As described by *Chialvo* and *Millonas* in [3], the state of an individual ant can be expressed by its position *r*, and orientation $\theta$. Since the response at a given time is assumed to be independent of the previous history of the individual, it is sufficient to specify a transition probability from one place and orientation $(r,\theta)$ to the next $(r^*,\theta^*)$ an instant later. In previous works [15,16], transition rules were derived and generalized from noisy response functions, which in turn were found to reproduce a number of experimental results with real ants. The response function can effectively be translated into a two-parameter transition rule between the cells by use of a pheromone weigthing function:

$$W(\sigma) = \left(1 + \frac{\sigma}{1+\gamma\sigma}\right)^\beta \quad (3.1)$$

This equation measures the relative probabilities of moving to a cite *r* (in our context, to a pixel) with pheromone density $\sigma(r)$. The parameter $\beta$ is associated with the osmotropotaxic sensitivity, recognised by *Wilson* as one of two fundamental different types of ants sense-data processing. *Osmotropotaxis*, is related to a kind of instantaneous pheromonal gradient following, while the other, *klinotaxis*, to a sequential method (though only the former will be considered in the present work as in [3]). Also it can be seen as a physiological inverse-noise parameter or gain. In practical terms, this parameter controls the degree of randomness with wich each ant follows the gradient of pheromone. On the other hand, $1/\gamma$ is the sensory capacity, which describes the fact that each ant's ability to sense pheromone decreases somewhat at high concentrations. In addition to the former equation, there is a weigthing factor $w(\Delta\theta)$, where $\Delta\theta$ is the change in direction at each time step, i.e. measures the magnitude of the difference in orientation. As an additional condition, each individual leaves a constant amount $\eta$ of pheromone at the pixel in which it is located at every time step *t*. This pheromone decays at each time step at a rate *k*. Then, the normalised transition probabilities on the lattice to go from cell *k* to cell *i* are given by $P_{ik}$ (in, [3]):

$$P_{ik} = \frac{W(\sigma_i)w(\Delta_i)}{\sum_{j/k}W(\sigma_j)w(\Delta_j)} \quad (3.2)$$

where the notation *j/k* indicates the sum over all the pixels *j* which are in the local neighbourhood of *k*. $\Delta_i$ measures the magnitude of the difference in orientation for the previous direction at time *t*-1. That is, since we use a neighbourhood composed of the cell and its eight neighbours, $\Delta_i$ can take the discrete values 0 through 4, and it is sufficient to assign a value $w_i$ for each of these changes of direction. *Chialvo et al* used the weights of $w_0$ =1 (same direction), $w_1$ =1/2, $w_2$ =1/4, $w_3$ =1/12 and $w_4$ =1/20 (U-turn). In addition, coherent results were found for $\eta$=0.07 (pheromone deposition rate), *k*=0.015 (pheromone evaporation rate), $\beta$=3.5 (osmotropotaxic sensitivity) and $\gamma$=0.2 (inverse of sensory capacity), where the emergence of well defined networks of trails were possible. For a detailed mathematical discussion of this model, and other conditions readers are reported to [3]. Except when indicated, these values will remain in the following framework. As an additional condition, each individual leaves a constant amount $\eta$ of pheromone at the pixel in which it is located at every time step t. Simultaneously, the pheromone evaporates at rate *k*, i.e., the pheromonal field will contain information about past movements of the organisms, but not arbitrarily in the past, since the field *forgets* its distant history due to evaporation in a time $\tau \cong 1/k$. As in [3], toroidal boundary conditions are imposed on the lattice to remove, as far as possible any boundary effects (e.g. one ant going out of the image at the south-west corner, will probably come in at the north-east corner).

In order to achieve emergent and *autocatalytic* mass behaviours around item groups on the *habitat*, instead of a constant pheromone deposition rate $\eta$ used in [6], a term not constant is included. This upgrade can significantly change the expected ant colony cognitive map (pheromonal field). The strategy follows an idea implemented by *Ramos* et al. [20], while extending the *Chialvo* model into digital image habitats. In here, however, this term should naturally be related with the amount of items in one specific region. So for instance, if we use $\Delta_h$ as that measure (i.e., the number of items present in one neighborhood), the pheromone deposition rate *T* for a specific ant at that specific cell (at time *t*), should change to a dynamic value (*p* is a constant = 0.0025): $T = \eta + p\Delta_h$. Notice that, if no objects are present, results expected by this extended model will be equal to those found by *Chialvo* and *Millonas* in [3], since $\Delta_h$ equals to zero. In this case, this is equivalent to say that only the swarm pheromonal field is affecting each ant choices, and not the *environment* - i.e. the expected network of trails depends largely on the initial random position of the colony, and in clusters formed in the initial configurations of pheromone, through relative distances. On the other hand, if this environmental term is added, a stable configuration will appear, which is largely independent on the initial conditions of the colony, and becomes more dependent on the nature of *items* itself.

## IV. STRESSING THE ROLE OF RESPONSE THRESHOLDS TO TASK-ASSOCIATED STIMULUS INTENSITIES

In order to model the behavior of ants associated to different tasks, as dropping and picking up objects, we suggest the use of combinations of different response thresholds. As we have seen before, there are two major factors that should influence any local action taken by the ant-like agent: the number of objects in his neighborhood, and their similarity (including the hypothetical object carried by one ant). *Lumer* and *Faieta* [14], use an average similarity (Eq. 2.3, section 2), mixing distances between objects with their number, incorporating it simultaneously into a response threshold function like the one of *Denebourg*'s (Eq. 2.1, 2.2, section II). Instead, in the present proposal, we suggest the use of combinations of two independent response threshold functions, each associated with a different environmental factor (or, stimuli intensity), that is, the number of objects in the area, and their similarity. Moreover, the computation of average similarities are avoided in the present algorithm, since this strategy can be somehow blind to the number of objects present in one specific neighborhood. In fact, in *Lumer* and *Faieta*'s work [14], there is an hypothetical chance of having the same average similarity value, respectively having one or, more objects present in that region. But, experimental evidences and observation in some types of ant colonies, can provide us with a different answer.

After *Wilson* [23], it is known that minors and majors in the polymorphic species of ants *Genus Pheidole*, have different response thresholds to task-associated stimulus intensities (i.e., division of labor). Recently, and inspired by this experimental evidence, *Bonabeau* et al. ([24,25]), proposed a family of response threshold functions in order to model this behavior. According to it, every individual has a response threshold for every task. Individuals engage in task performance when the level of the task-associated stimuli exceeds their thresholds. Authors defined $s$ as the intensity of a stimulus associated with a particular task, i.e. $s$ can be a number of encounters, a chemical concentration, or any quantitative cue sensed by individuals. A response threshold $\theta$, expressed in units of stimulus intensity, is an internal variable that determines the tendency of an individual to respond to the stimulus $s$ and perform the associated task. More precisely, $\theta$ is such that the probability of response is low for $s \ll \theta$ and high for $s \gg \theta$. One family of response functions $T_\theta(s)$ (the probability of performing the task as a function of stimulus intensity $s$), that satisfy this requirement is given by (*Bonabeau* et al., [24,25]):

$$T_\theta(s) = \frac{s^n}{s^n + \theta^n} \quad (4.1)$$

where $n>1$ determines the steepness of the threshold (normally $n=2$, but similar results can be obtained with other values of $n>1$). Now, at $s = \theta$, this probability is exactly ½. Therefore, individuals with a lower value of $\theta$ are likely to respond to a lower level of stimulus. In order to take account on the number of objects present in one neighborhood, Eq. 4.1, was used (where, $n$ now stands for the number of objects present in one neighborhood, and $\theta = 5$), defining χ (Eq. 4.2) as the response threshold associated to the number of items present in a 3 x 3 region around $r$ (one specific grid location):

$$\chi = \frac{n^2}{n^2 + 5^2} \quad (4.2)$$

$$\delta = \left(\frac{k_1}{k_1 + d}\right)^2 \quad (4.3)$$

$$\varepsilon = \left(\frac{d}{k_2 + d}\right)^2 \quad (4.4)$$

Now, in order to take account on the hypothetical similarity between objects, and in each ant action due to this factor, a Euclidean normalized distance $d$ is computed within all the pairs of objects present in that 3 x 3 region around $r$. Being $a$ and $b$, a pair of objects, and $f_a(i)$, $f_b(i)$ their respective feature vectors (being each object defined by $F$ features), then $d = (1/d_{max}).[(1/F).\Sigma_{i=1,F}(f_a(i)-f_b(i))^2]^{1/2}$. Clearly, this distance $d$ reaches its maximum (=1, since $d$ is normalized by $d_{max}$) when two objects are maximally different, and $d=0$ when they are equally defined by the same $F$ features. Then, $\delta$ and $\varepsilon$ (Eqs. 4.3, 4.4), are respectively defined as the response threshold functions associated to the similarity of objects, in case of dropping an object (Eq. 4.3), and picking it up (Eq. 4.4), at site $r$. Note that these functions are similar to those proposed by *Denebourg* et al. [7] ($k_1$ and $k_2$, are threshold constants), while defining probabilities for picking up or to deposit an item (Eqs. 2.1, 2.2, section II). In here, however, we use them in reversed order, substituting $f$ by $d$ (where $f$ represented, for *Denebourg* et al., the perceived fraction of items in the neighborhood of one agent, having in mind a robotic implementation, which is not the case in here). Let us now review the behavior of one of these functions. The probability $\delta$ for a specific moving loaded agent to deposit an item at site $r$, is given by Eq. 4.3. When $d \ll k_1$ (i.e., $d$ close to 0), $\delta$ is close to 1, that is, the probability of dropping an item is high when the similarity between the loaded object and one present in the region around $r$, is high. Similarly, the probability $\delta$ for a specific moving loaded agent to deposit an item at site $r$, is low, when $d \gg k_1$ (i.e., $d$ close to 1), $\delta$ is close to 0, that is, items are unlikely to be deposited together since they are very different. Now, in order to deal and represent different stimulus intensities (number of items and their similarity), the strategy uses a composition of the above defined response threshold functions (Eq. 4.2, 4.3 and 4.4). These composed functions are used in every action taken by an agent, present at each visited site in the environment.

TYPES OF HYBRID RESPONSE FUNCTIONS USED

| Function Types | Picking Probability | Dropping Probability |
|---|---|---|
| #1 | $P_p = (1-\chi).\varepsilon$ | $P_d = \chi.\delta$ |
| #2 | (a) $P_p = (1-\chi).\varepsilon$ <br> (b) $P_p = \varepsilon$ | (a) $P_d = \chi.\delta$ <br> (b) $P_d = \delta$ |
| #3 | (a) $P_p = 1-\chi$ <br> (b) $P_p = \varepsilon$ | (a) $P_d = \chi$ <br> (b) $P_d = \delta$ |
| #4 | *Lumer & Faieta* Eq.(2.4) | *Lumer & Faieta* Eq.(2.5) |

Table 1 – Types of picking ($P_p$) and dropping ($P_d$) probability functions used for several tests. In #2,3 half of the ants used one probability function (*a*), while the rest used the other function (*b*). In #4, the LF algorithm (section II) was fully implemented and followed, but using a toroidal grid.

These composed probabilities are resumed in table 1, and were used as test functions in the "4 classes x 200 Gaussian distributed points" problem (fig. 2) proposed by *Lumer* and *Faieta* [14] (section II), in order to illustrate the functioning of the algorithm. On the other hand, to evaluate the algorithm behavior, a simple entropy definition is proposed. For a finite number of *n* type *A* items, placed into a finite area grid, the entropy of *A* type objects can be defined as the normalized sum, over all *n*, of the number of empty cells *e* (or occupied by objects different from *A*), surrounding each item *A* ($e_{max}$ = 8, in 3 x 3 regions), that is, $E_A = (\sum e_i) / (n . e_{max})$. As its obvious, several configurations lead to different values of entropy, where $E_A$ reaches its maximal value ($E_A$ = 1) when all type *A* items are disconnected from each other. Disconnected clusters of type *A* items, lead also to an increase in the value of entropy.

## V. RESULTS ON A "4 CLASSES X 200 GAUSSIAN DISTRIBUTED POINTS" PROBLEM

As mentioned before, we decide to test the algorithm using the same problem as *Lumer* and *Faieta*, introduced by them in [14]. This problem consists of 800 points, represented by two features each. That is, the attribute space is $R^2$, and the values of the two attributes for each object correspond to its coordinates (*x,y*) in $R^2$. Four clusters of 200 points each were then generated in attribute space, with *x* and *y* distributed according to Normal (or Gaussian) distributions $N(\mu,\sigma)$ of average $\mu$ and variance $\sigma^2$ - see figure 2 for details. The 800 data points (items) were then assigned at random locations on a 57 x 57 non-parametric toroidal grid, and the clustering algorithm was run with 80 ants, using the function types specified in table 1. Generally, the following empirical rules were followed, since they lead to good results: $A=4.n_o$, $n_a=A/40$, and $n_a/n_o=0.1$, where *A*

**Algorithm.**
High-level description of the *ACLUSTER* algorithm proposed

```
/* Initialization */
For every item o_i do
    Place o_i randomly on grid
End For
For all agents do
    Place agent at randomly selected site
End For
/* Main loop */
For t = 1 to t_max do
    For all agents do
        sum = 0
        Count the number of items n around site r
        If ((agent unladen) and (site r occupied by item o_i)) then
            For all sites around r with items present do
                /* According to Eqs. 4.2, 4.4 and Table 1 (section 4) */
                Compute d,χ, ε and P_p
                Draw random real number R between 0 and 1
                If (R ≤ P_p) then
                    sum = sum + 1
                End If
            End For
            If ((sum ≥ n/2) or (n = 0)) then
                Pick up item o_i
            End If
        Else If ((agent carrying item o_i) and (site r empty)) then
            For all sites around r with items present do
                /* According to Eqs. 4.2, 4.3 and Table 1 (section 4) */
                Compute d,χ, δ and P_d
                Draw random real number R between 0 and 1
                If (R ≤ P_d) then
                    sum = sum + 1
                End If
            End For
            If (sum ≥ n/2) then
                Drop item o_i
            End If
        End If
        /* According to Eqs. 3.1 and 3.2 (section 3) */
        Compute W(σ) and P_ik
        Move to a selected neighboring site not occupied by other agent
        Count the number of items n around that new site r
        Increase pheromone at site r according to n, that is:
            P_r = P_r+[η+(n/α)]
    End For
    Evaporate pheromone by K, at all grid sites
End For
Print location of items
/* Values of parameters used in experiments */
k_1 = 0.1, k_2 = 0.3, K = 0.015, η = 0.07, α = 400,
β=3.5, γ=0.2, t_max = 10^6 steps.
```

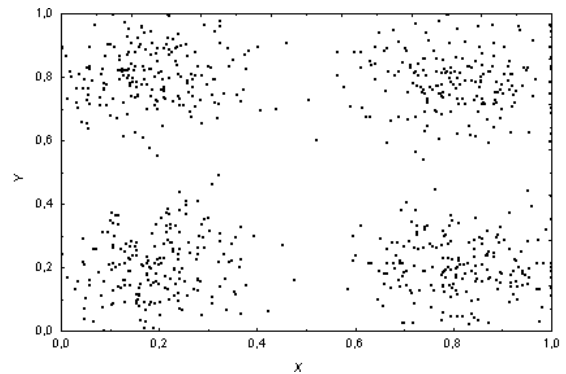

Fig.2. Distribution of points in "attribute space": 4 clusters of 200 points each were generated in attribute space, with *x* and *y* distributed according to Normal (or Gaussian) distributions $N(\mu,\sigma)$: $A \equiv [x \approx N(0.2,0.1), y \approx N(0.2,0.1)]$, $B \equiv [x \approx N(0.8,0.1), y \approx N(0.2,0.1)]$, $C \equiv [x \approx N(0.8,0.1), y \approx N(0.8,0.1)]$, $D \equiv [x \approx N(0.2,0.1), y \approx N(0.8,0.1)]$, for objects type *A*, *B*, *C* and *D*, respectively.

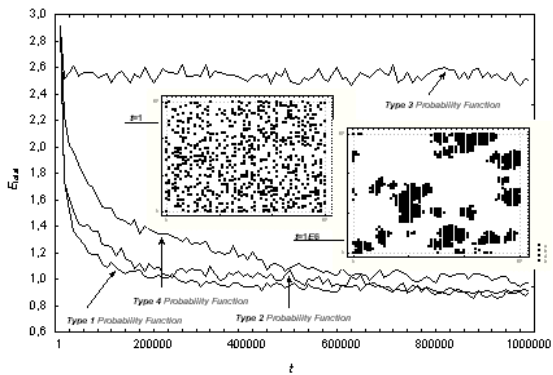

Fig.3. Total entropy, $E_{total} = E_a + E_b + E_c + E_d$, in time, as the swarm evolves new solutions in clustering four type of objects. Four graphs are shown which correspond to four types of Probability functions (dropping and picking) analyzed (see table 1).

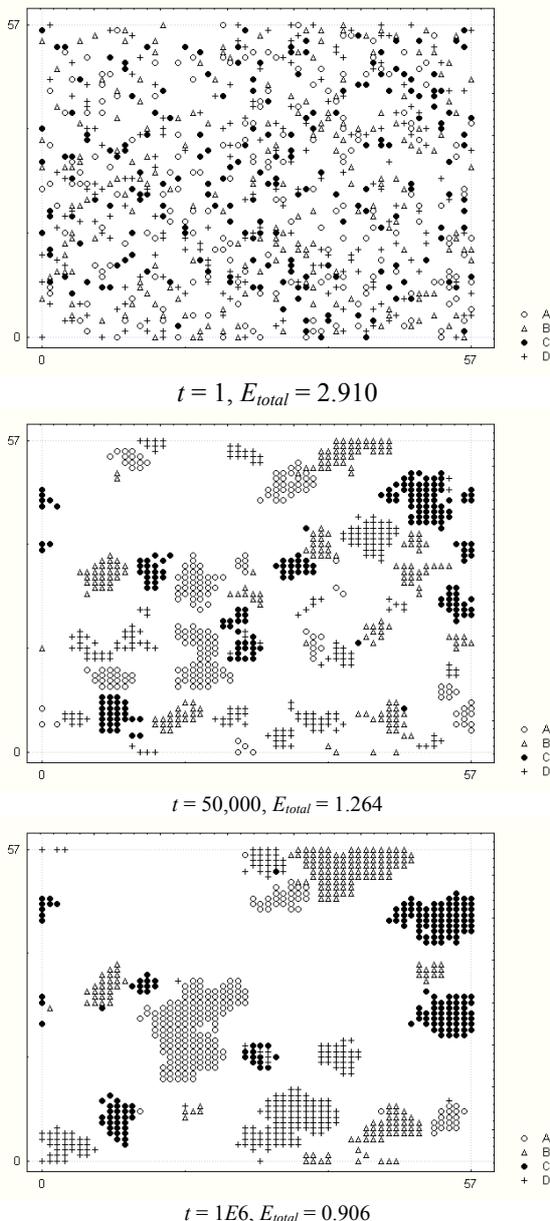

Fig. 4. Spatial distribution of 800 items on a 57 x 57 non-parametric toroidal grid at several time steps. At $t=1$, four types of items are randomly allocated into the grid. As time evolves, several homogenous clusters emerge due to the ant colony action, and as expected the total entropy decreases. In order to illustrate the behavior of the algorithm, items that belong to different clusters (see fig. 2), were represented by different symbols: o, $\Delta$, • and +. Type 1 probability function was used with $k_1$=0.1 and $k_2$=0.3.

is the grid area, $n_o$ is the number of objects, and $n_a$ the number of ants used. As a final result, objects that are clustered together belong generally to the same initial distribution, and objects that do not belong to the same initial distribution are found generally in different clusters. In figure 6.1.b, the evolution of total entropy ($E_{total}=E_A+E_B+E_C+E_D$), for $10^6$ iterations (as those used in [14]) was plotted for four different type functions. It is clear to see that probabilistic functions type #3, are the worse in terms of clustering the different items, while the rest (including the algorithm proposed by *Lumer* and *Faieta* [14]) have similar behaviors, and indeed reduce drastically the value of entropy of those configurations. We can also get an idea of how the new algorithm clusters the different items, while the algorithm proceeds (fig. 4). In this case, type function #1 was used. This gives an description of how, initially randomly deposited items at $t$=1, are spatially distributed according to their similarities, by the proposed algorithm. Finally, note that a toroidal grid was used.

## VI. LATENT SEMANTIC ANALYSIS AND DOCUMENT FILTERING: PRELIMINARY RESULTS ON NEWSPAPERS

A major challenge in cyberspace in general, or in document filtering specifically, is to automate the delivery of relevant information to individual users. In the last few years, several systems have tried to cope with information filtering in several ways, like collaborative filtering (*webrings*), tracking the user behaviour while he reads (*anatagonomy*), using genetic algorithms to evolve profiles, using ontologies to represent page content and user interest, but current results are not very reliable. Other approaches use Self Organizing Maps (SOM, *Kohonen*), as in *Honkela* and *Kohonen*'s WEBSOM [13].

The lack of accurate retrieval in information filtering may due to the lexical matching algorithm used in most of them. For instance, synonymy and polysemy are very important factors, but it seems that most systems does not consider this fact [18]. One of the algorithms that has been showed a good accuracy is Latent semantic analysis (LSA, [6]), based on the assumption that meaning of words can be represented in a multidimensional space. This algorithm has been recently and succesfully extended into *LSAmercury*, an information filtering engine, by *Quesada*, *Merelo* et al. [18], allowing users to edit their own profiles. Profiles, queries and documents are all represented as LSA vectors, which add flexibility to the operations that the user can perform on her profile, while using simultaneously non-textual matching and previous actions informations to infer user preferences.

The present work uses LSA as a feature extraction method, in order to map 931 words of an article at a Spanish newspaper. In the LSA model [6, 13], terms and documents are represented by an $m$ x $n$ incidence matrix $A$. Each of the $n_i$ unique terms in the document collection are assigned a row in the matrix, while each document is assigned a column. SVD is applied to the resulting matrix, and the main "axes" are them obtained. Words are projected onto those axes, resulting similar vector values for words with a similar meaning.

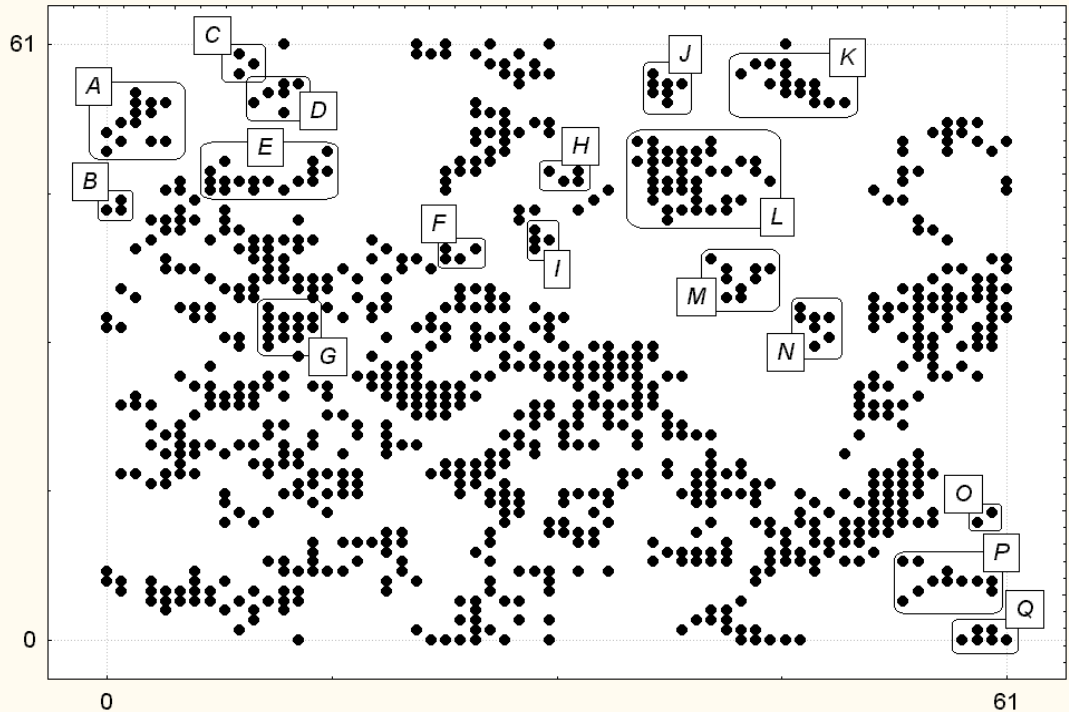

Fig. 5 – Spatial distribution of 931 items (words taken from an article at a Spanish newspaper) on a 61 x 61 non-parametric toroidal grid, at $t=10^6$. 91 ants used type 2 probability response functions, with $k_1=0.1$ and $k_2=0.3$. Some independent clusters examples are: *(A) anunció, bilbao, embargo, titulos, entre, hacer, necesídad, tras, vida, lider, cualquier, derechos, medida.(B) dirigentes, prensa, ciu. (C) discos, amigos, grandes. (D) hechos, piloto, miedo, tipo, cd, informes. (E) difícil, gobierno, justicia, crisis, voluntad, creó, elección, horas, frente, técnica, unas, tarde, familia, sargento, necesídad, red, obra. (F) voz, puenlo, papel, asseguró. (G) nuestro, europea, china, ahora, poder, hasta, mucho, compañía, nacionalistas, cambio, asesinado, autor, nuevo.(H) rodríguez, vez, tramitación, gran. (I) se, declara, junto, administración.(J) final, visita, cataluña, puerta, final, jurisprudencia, todas.(K) fallo, ejército, bajo, real, situación, mission, liga, teatro, decision, queda, nacionalismo, pasado, director, plan, manos. (L) unica, blancos, ibarra, intensidad, nuevas, las, persona, parlamento, españoles, tarde, seis, otros, euro, elecciones, servicios, podría, otra, tiene, nada, posibilidad, hablar, porque, música, puntos, compromiso, dentro, doctrina, fiscal, abc, derecho, atentado, sistema, carrera, razón, televisión, semanas, mundo, natural, mitad. (M) mayo, parís, ciento, consejo, reconoció, me, pero, lo, ocasión. (N) incluso, pnv, luis, momentos, miembros, regimen, ee.(O) cabeza, ex.(P) oea, municipals, mujer, ayuntamiento, cosas, toda, novedades, debate, firmado. (Q) domidomingo, estado, otro, primeros, estamos, no.*

Thus, each word uses a 50 feature vector. Since we had 931 items (words) to self-organize by the swarm, 91 ants were used (see section V), on a 61 x 61 non-parametric toroidal grid. Figure 3 shows the final result at $t=10^6$.

## VII. CONCLUSIONS

We have presented in this paper a new ant-based algorithm named ACLUSTER for data unsupervised clustering and data exploratory analysis. The aim of this paper was to draw a clear parallel between a mode of problem-solving in social insects and a distributed, reactive, algorithmic approach. Some of the mechanisms underlying corpse clustering, brood sorting and those that can explain the worker's behavioral flexiblity, as regulation of labor and allocation of tasks have first been analysed. The role of response thresholds to task-associated stimulus intensities was then stressed as an important part of the strategy, and were in fact incorporated into the algorithm by using compositions of different response functions. These compositions allows the strategy not only to be more accurate relatively to behaviours found in nature as avoids short-term memory based strategies, as well as the use of several artificial ant types (using different speeds), present in some recent approaches. Behavioral switches as used by *Lumer* and *Faieta* [14], were also avoided, in order to maintain simplicity and to avoid complex parameter settings to be performed by the domain expert. At the level of agent moves in the grid, a truly stigmergic model was adopted and extended in order to deal with clusters of objects, avoiding randomly moves which can be counterproductive in the distributed search performed by the swarm, and adopted by many past models. Results speak for themselves. While achieving similar results compared to *Lumer*'s and *Faieta* model [14], as pointed by the spatial entropy of solutions at each time iteration, the present algorithm is by far more simple. Moreover, for some of response thresholds compositions used, results are better while using the present algorithm for the majority of time iterations, that is, entropy is always lower, even if at the end they tend to the same value. As a final advantage, ACLUSTER does not require any initial information about the future classification, such as an initial partition or an initial number of classes. This novel strategy was then applied for the first time to document word clustering. While maintaining some coherence for some words and for some clusters, results, however, are far from being optimal. This probably occurred due to the relatively quality of LSA features regarding this specific newspaper article. At this moment of the research ACLUSTER has not been extensively tested within this type of features, and our conclusions are limited because of that.